\documentclass{article}

\usepackage[main, final]{neurips_2026}

\usepackage[utf8]{inputenc} 
\usepackage[T1]{fontenc}    
\usepackage{hyperref}       
\usepackage{url}            
\usepackage{booktabs}       
\usepackage{amsfonts}       
\usepackage{nicefrac}       
\usepackage{microtype}      
\usepackage{xcolor}         

\usepackage{tabularx}
\usepackage{multirow}
\usepackage{graphicx}
\usepackage[table]{xcolor}
\usepackage{amsmath}
\usepackage{placeins}
\usepackage{subfigure}
\usepackage{caption}
\definecolor{OursBlue}{RGB}{232, 244, 255} 
\definecolor{DeltaGreen}{RGB}{0, 150, 0} 
\newcommand{\ours}{CoLVR}

\hypersetup{
    colorlinks=true,
    linkcolor=red,
    citecolor=blue, 
    urlcolor=green
}

\title{\ours: Enhancing Exploratory Latent Visual Reasoning via Contrastive Optimization}

\author{
    Ziyang Ding$^{1,2}$\thanks{This work was done during his internship at Shanghai Artificial Intelligence Laboratory.} ,
    Linjian Meng$^{2}$,
    Yiming Wu$^{2}$,
    Yuhan Li$^{2,3}$,
    Yuhao Liu$^{4}$,
    Zhen Zhao$^{2}$\thanks{Corresponding author.}\\
    $^{1}$Shanghai Jiao Tong University, 
    $^{2}$Shanghai Artificial Intelligence  Laboratory \\
    $^{3}$Nanjing University, 
    $^{4}$Shandong University \\
    \texttt{zhen.zhao@outlook.com} \\
}

\begin{document}

\maketitle

\begin{abstract}

  Due to the potential for exploratory reasoning of Latent Visual Reasoning, recent works tend to enable MLLMs (Multimodal Large Language Models) to perform visual reasoning by propagating continuous hidden states instead of decoding intermediate steps into discrete tokens. However, existing works typically rely on hard alignment objectives to force latent representations to match predefined visual features, thereby severely limiting the exploratory of latent reasoning process. To address this problem, we propose \textbf{\ours} (\textbf{C}ontrastive \textbf{O}ptimization for \textbf{L}atent \textbf{V}isual \textbf{R}easoning). To obtain a more exploratory visual reasoning, CoLVR introduces a latent contrastive training framework. Firstly, CoLVR learns diverse and exploratory representations with a latent contrastive objective guided by angle-based perturbation, which expands the semantic latent space and avoids over-constrained embedding. Then, CoLVR employs a latent trajectory contrastive reward for RL (Reinforcement Learning) post-training to enable fine-grained optimization of latent visual reasoning process and thus fostering diverse reasoning behaviors. Experiments demonstrate that \ours{} significantly enhances the exploratory capability of latent representations, achieving average improvements of \textbf{5.83\%} on VSP and \textbf{8.00\%} on Jigsaw, while also outperforming existing latent models on out of domain benchmarks, with a \textbf{3.40\%} gain on MMStar. The data, codes, and models are released at \url{https://github.com/Oscar-dzy/CoLVR}.
\end{abstract}

\section{Introduction}
\label{section:intro}
Recent progress on VQA (Visual Question Answering) tasks in MLLMs (Multimodal Large Language Models) is significant~\cite{kimiteam2025kimivltechnicalreport,openai2024gpt4technicalreport,bai2025qwen25vltechnicalreport,chen2024internvl,zhao2025cot}. These methods construct multimodal chains of thought with rich visual information through image editing and generation, which improves visual exploration and understanding~\cite{he2025diffthinkergenerativemultimodalreasoning,xu2026visual,zheng2026deepeyes}. However, they rely on autoregressive generation over discrete text tokens, which leads to long reasoning sequences and high latency. More importantly, discrete tokens are inherently limited in representing fine-grained and continuous visual features, which are essential for many visual reasoning tasks~\cite{yang2025machine}.

\begin{figure}
    \centering
    
    \subfigure[Latent Token Exploratory Comparison.]{
        \centering
        \includegraphics[width=\linewidth]{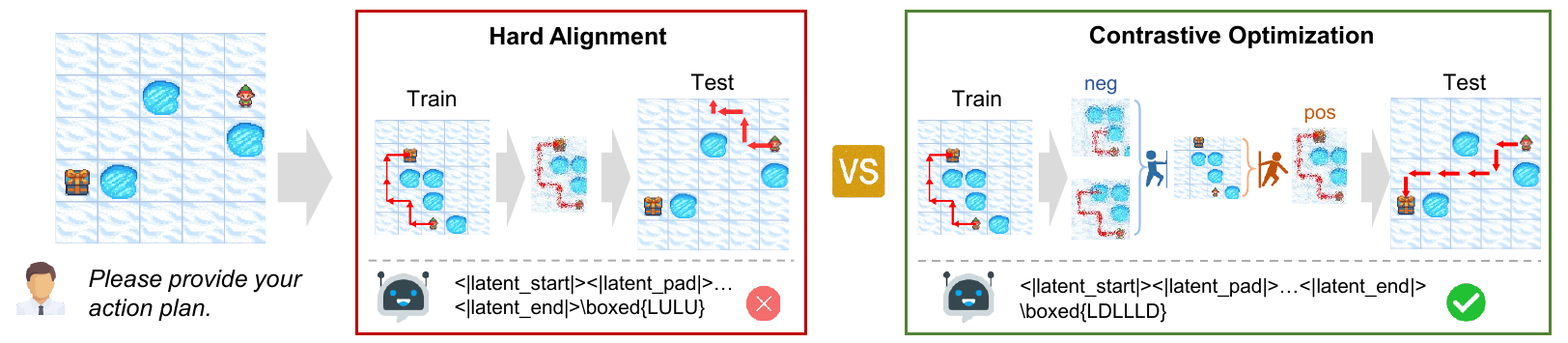}
        \label{fig:caption_a}
    }

    \subfigure[UMAP Visualization of Latent Tokens.]{
        \centering
        \includegraphics[width=\linewidth]{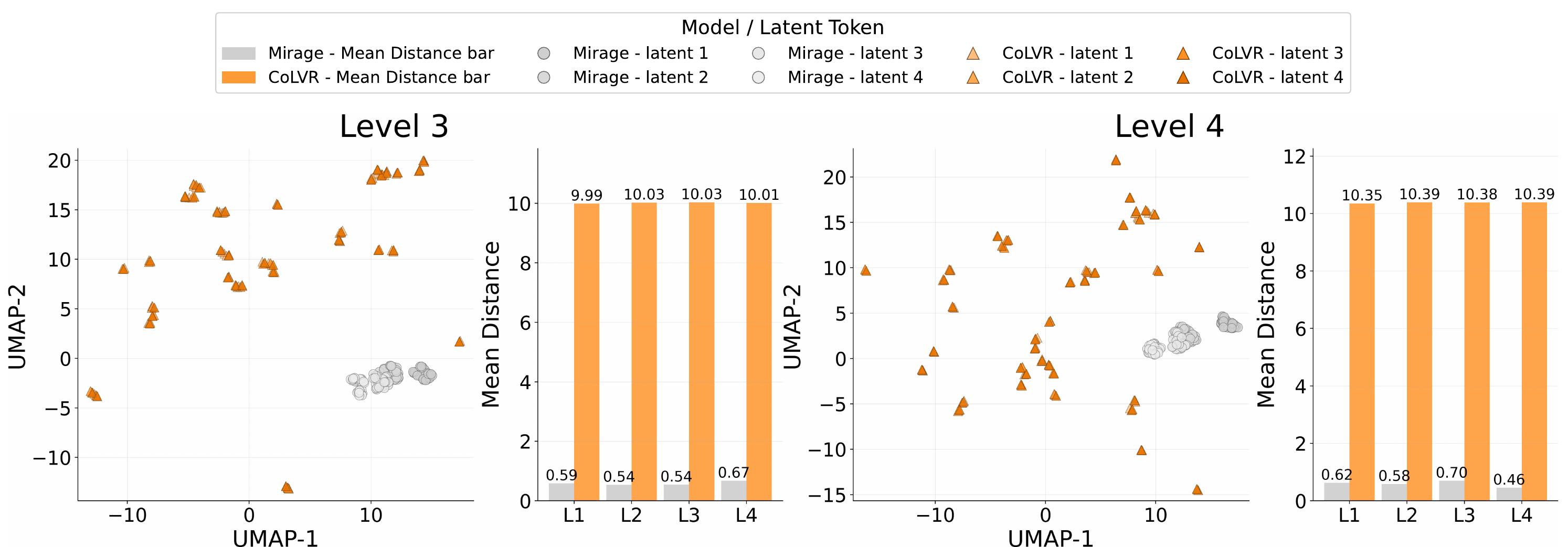}
        \label{fig:caption_b}
    }
    
    \caption{(a) \textbf{Comparison of Latent Token Exploratory Capacity Under Hard Alignment and Contrastive Optimization}. Hard alignment enforces rigid feature matching and leads to fixed reasoning paths, while contrastive optimization encourages more flexible latent trajectories. (b) \textbf{UMAP-based 2D visualization of the first four latent tokens from Mirage and \ours{}}, where hidden states are projected into a two-dimensional space and cluster compactness is quantified via the mean distance to centroids. The results highlight that \ours{} learns more exploratory latent representations.}
    \label{fig:caption}

\end{figure}

To address these limitations, recent studies introduce \textbf{Latent Visual Reasoning}~\cite{yang2025machine,ma2025cocova,wang2025monetreasoninglatentvisual}, where intermediate reasoning steps are no longer decoded into text but propagated as continuous hidden states. This paradigm enables models to reason directly in the latent space, leading to more efficient and more exploratory processing of visual information. Despite its promise, existing approaches still lacks the flexibility since they often directly force latent states to match predefined visual representations, such as particular image patches or object-level regions, through hard alignment objectives, as shown in Figure~\ref{fig:caption_a}.





Such hard alignment further drives latent tokens into a homogeneous latent space, limiting their exploratory capability~\cite{wang2026foresttreeslatentsuperposition}. To empirically verify this effect, we analyze a classical Latent Visual Reasoning approach called Mirage~\cite{yang2025machine}, which explicitly aligns latent tokens with grouped image patches. By projecting latent tokens from different test samples into a 2D space using UMAP~\cite{mcinnes2020umapuniformmanifoldapproximation} as shown in Figure~\ref{fig:caption_b}, we observe that Mirage latent tokens from the same reasoning step collapse into highly concentrated clusters, with an average distance to the cluster centroids of only 0.5 $\sim$ 0.7. This provides empirical evidence that hard alignments reduce the diversity of latent representations, implying a decline on the weaken exploratory reasoning. To mitigate this issue, recent work such as Laser~\cite{wang2026foresttreeslatentsuperposition} aligns latent tokens with future visual reasoning states. However, it still relies on predefined alignment targets, which implicitly constrain the latent semantic space, thereby only partially alleviating the limited flexibility and exploratory capacity of latent visual reasoning. 

Furthermore, recent reinforcement learning methods for latent reasoning, such as VLPO~\cite{wang2025monetreasoninglatentvisual}, attempt to optimize latent tokens by modeling their probability distributions. However, these approaches still adopt predefined distribution assumptions and trajectory-level alignment objectives, forcing latent trajectories to match specific latent states observed during rollout. As a result, such optimization remains centered on matching predefined patterns instead of encouraging exploratory latent reasoning.

To address these challenges, we propose \textbf{\ours}{} (\textbf{C}ontrastive \textbf{O}ptimization for \textbf{L}atent \textbf{V}isual \textbf{R}easoning), a latent contrastive training framework toward more flexible and exploratory visual reasoning. In contrast to hard alignment methods that constrain latent visual tokens toward predefined patches, regions, or rollout states, \ours{} does not specify an explicit target pattern for each reasoning step. Instead, \ours{} optimizes relative contrastive relationships among latent visual states, thereby preserving the freedom of latent exploration while still providing task-relevant supervision. Specifically, after a warm-up stage, we construct positive and negative samples via angle-based perturbations and introduce a contrastive objective in the latent space. Positive samples consist of image data paired only with result hints, while negative samples are generated by applying angular perturbations to latent representations. This design reduces the reliance on process-level visual annotations and encourages the model to learn more discriminative and diverse latent representations. Furthermore, we extend the contrastive principle to the reinforcement learning stage by introducing a latent-trajectory contrastive reward over hidden states, which provides relative preference signals among latent trajectories rather than enforcing predefined trajectory targets. This provides latent-level preference signals that encourage task-relevant latent rollouts during reinforcement learning. 

Extensive experiments demonstrate the effectiveness of our method. In particular, \ours{} achieves an average improvement of \textbf{5.83\%} on the VSP (Visual Spatial Planning) task and \textbf{8.00\%} in Jigsaw task. It also significantly outperforming constraint-based latent models across multiple OOD benchmarks, including a \textbf{3.40\%} improvement on MMStar. These results highlight the strong exploratory capability of the latent tokens learned by \ours{}.

Our contributions are summarized as follows: \begin{enumerate} 
    \item We identify and analyze a fundamental limitation of existing latent visual reasoning methods: hard alignment objectives constrain latent representations into a homogeneous latent space, thereby limiting the exploratory capability of latent tokens.
    \item We propose \ours, a latent visual reasoning training framework that introduces latent contrastive objectives in both the fine-tuning and reinforcement learning stages, in order to enhance the exploratory capabilities of latent tokens.
    \item Extensive experiments demonstrate that our method effectively improves the exploratory behavior and flexibility of the model, thereby leading to better performance on visual understanding tasks.
\end{enumerate}

\section{Related Work}
\subsection{Latent Visual Reasoning}
Latent reasoning was pioneered in the Coconut~\cite{haotraining} framework, which eliminates the need for explicit decoding during intermediate steps. Rather than generating discrete tokens, the model propagates hidden states as a continuous reasoning medium, using the final state of one step as the input embedding for the next. Recent work has extended this paradigm to multimodal large language models to facilitate visual reasoning within the latent space. Mirage~\cite{yang2025machine} aligns image patches with hidden states to integrate visual context, while LVR~\cite{li2026latent} employs region-of-interest indexing to link latent states with localized image features. Similarly, Chain-of-Visual-Thought~\cite{qin2025chain} enhances fine-grained understanding by aligning hidden states with structured signals like segmentation and depth maps. However, these approaches rely on explicit alignment with preprocessed features, which constrains the latent space and often restricts latent tokens to a homogeneous feature set, ultimately limiting the exploration of diverse reasoning trajectories.

\subsection{Contrastive Learning}
Contrastive learning has been widely adopted in multimodal learning because it optimizes relative similarities by pulling positive samples closer while pushing negative samples apart in the feature space,  thereby enhancing the discriminability and flexibility of learned representations~\cite{su2025enhancing}.  This relative formulation provides a way to structure representation spaces without requiring each representation to match a fixed target feature. For example, HACL~\cite{jiang2024hallucination} constructs hallucinated captions as negative samples and applies contrastive learning to mitigate the modality gap between vision and language. MuCo~\cite{gu2026muco} improves multimodal embeddings by modeling contrastive relationships across different query-image pairs. CogFlow~\cite{chen2026cogflowbridgingperceptionreasoning} further incorporates contrastive principles into the reinforcement learning stage and introduces a knowledge initialization reward to guide optimization. However, the role of contrastive learning in optimizing latent reasoning trajectories and enhancing the exploratory capacity of latent tokens remains insufficiently explored. Our work investigates this direction by introducing latent contrastive objectives into both supervised fine-tuning and reinforcement learning, targeting latent representations and reasoning trajectories rather than multimodal embedding quality alone.
\section{Methods}

\subsection{Problem Formulation}
In this section, we present a unified formulation of latent visual reasoning. As shown in Figure~\ref{fig:framework}(a), given an input query $Q$ and an image $I$, the image is first encoded by a visual encoder $f_{enc}$ and projected into a high-dimensional semantic space. The resulting representation is then combined with the textual input and fed into a multimodal large language model $f_{\theta}$ to generate an output sequence that contains a latent reasoning sequence. To formalize this process, we introduce latent reasoning tokens, where \texttt{<|latent\_start|>} and \texttt{<|latent\_end|>} denote the beginning and end of the reasoning segment, and \texttt{<|latent\_pad|>} denotes intermediate latent tokens. The overall generation process is defined as:
\begin{equation}
    \bigl\{ \text{\small\texttt{<|latent\_start|>}}, 
    \underset{K}{\underbrace{\text{\small\texttt{<|latent\_pad|>}}}},\ 
    \text{\small\texttt{<|latent\_end|>}},\ \text{Answer} \bigr\}
    = f_{\theta}\bigl(Q,\ \text{Proj}(f_{\text{enc}}(I))\bigr).
\end{equation}
During latent reasoning, the model does not decode the last hidden state at intermediate steps. Instead, it directly feeds the hidden state as the input embedding for the next \texttt{<|latent\_pad|>}, enabling iterative propagation over $K$ steps in a continuous latent space. This mechanism allows the model to implicitly represent and reason over visual information without relying on explicit symbolic decoding.

\subsection{Warm Up Alignment}
\begin{figure}
  \centering
  \includegraphics[width=0.98\linewidth]{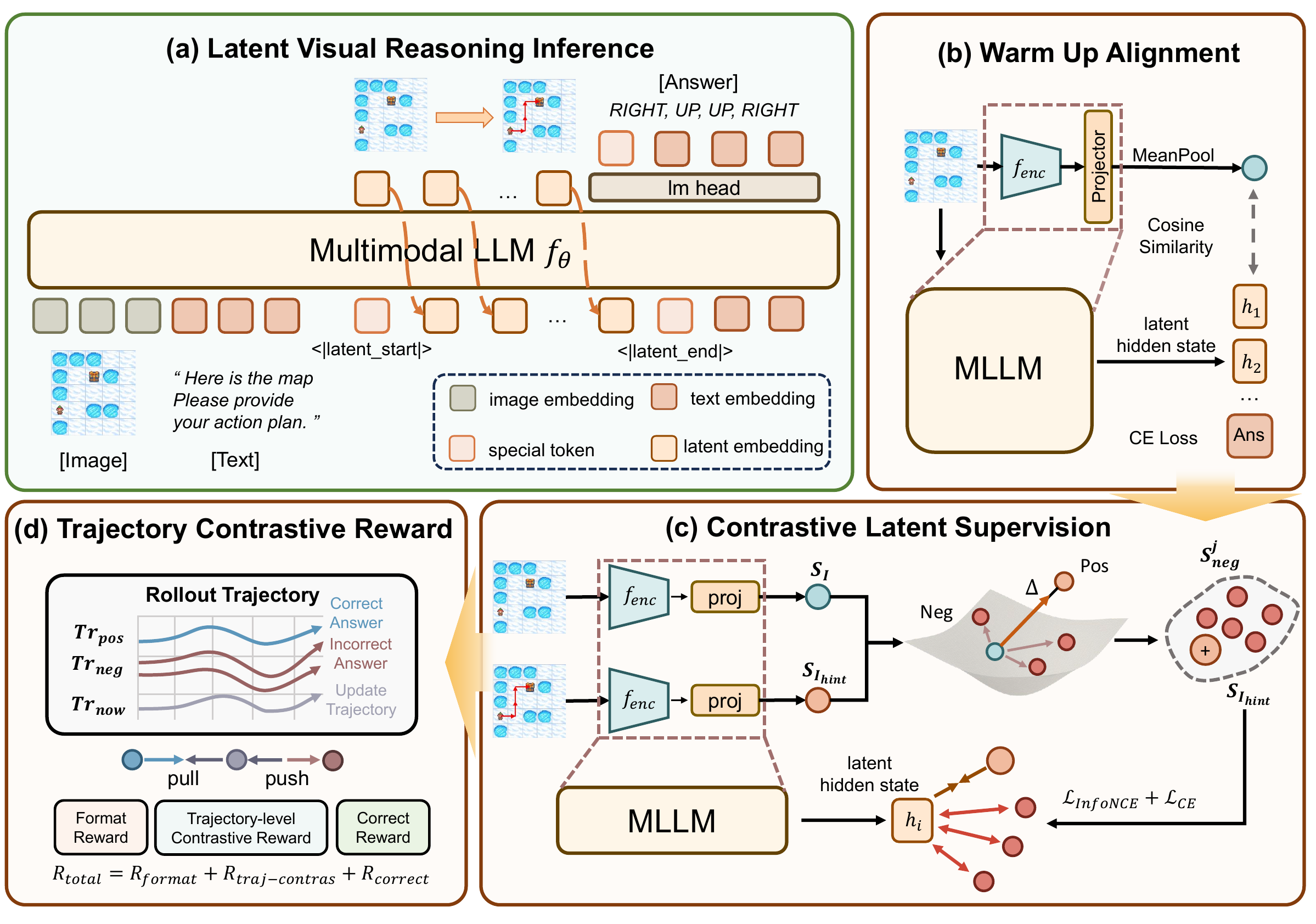}
  \caption{\textbf{The framework of \ours}. After a warm-up stage, \ours{} perform latent contrastive learning by constructing positive and negative samples via angular perturbations to encourage exploratory latent token representations. Additionally, we introduce a latent trajectory-based reward within the GRPO process to further optimize and sustain the exploration capability of latent tokens.}
  \label{fig:framework}
\end{figure}

In the initial training stage, we design a feature alignment based initialization strategy to enable the model to acquire preliminary reasoning capability in the visual latent space (as shown in Figure~\ref{fig:framework}(b)). Specifically, the input image is processed by a visual encoder $f_{\text{enc}}$ and a projector to obtain semantic features, followed by MeanPool(·) over the patch dimension to extract a global representation $S$:
\begin{equation}
    S = \mathrm{MeanPool}(\mathrm{Proj}(f_{\text{enc}}(I))).
\end{equation}
Based on this representation, we align the hidden states $h_i$ of all latent tokens with $S$ using cosine similarity, which provides an initial semantic constraint for latent reasoning. To preserve textual reasoning ability during alignment, we further introduce an auxiliary cross-entropy loss on the answer to stabilize language generation and reasoning performance:
\begin{equation}
    \mathcal{L}_{\text{warmup}} = \frac{1}{K} \sum_i \left(1 - \text{cos\_sim}(h_i, S)\right) + \lambda_1 \mathcal{L}_{\text{CE}}.
\end{equation}

\subsection{Latent Contrastive Supervised Fine-Tuning}
In the fine-tuning stage, we introduce a latent contrastive objective via angle-based perturbations to replace the commonly used hard alignment strategy (Figure~\ref{fig:framework}(c)). Specifically, we construct both positive and negative samples. The positive sample is defined as the augmented input image $I_{\text{hint}}$, which provides semantic guidance toward the correct answer and characterizes the feature distribution of the correct reasoning direction. The negative sample is designed to be structurally informative rather than random noisy perturbations. We aim to construct samples that are semantically close to the correct direction while remaining clearly deviated.


To this end, we follow the same feature extraction pipeline as in the warm up stage to obtain feature representations of the original input image and the image augmented with hints, denoted as $S_{I}$ and $S_{I_{\text{hint}}}$. Their difference defines a direction vector $\Delta = S_{I_{\text{hint}}} - S_{I}$, which represents a semantic trajectory from the original perception $S_{I}$ to the correct understanding $S_{I_{\text{hint}}}$. This trajectory represents the expected reasoning path in latent space. Accordingly, we view latent token learning as fitting this trajectory, where the model must identify and approach the direction that most effectively leads to the correct answer among multiple candidates.

Based on this trajectory, we construct structured perturbations to generate negative samples. We first sample a random perturbation vector $\epsilon$ from a standard normal distribution $\mathcal{N}(0, I)$ and compute its component orthogonal to the trajectory direction as
\begin{equation}
\eta = \epsilon - \frac{\epsilon \cdot \Delta}{\Delta^2} \cdot \Delta.
\end{equation}
We then normalize both the trajectory vector $\Delta_{\text{norm}}$ and the perturbation component $\eta_{\text{norm}}$, and rotate the trajectory by a randomly sampled angle $\theta_{\text{neg}}$ within a predefined range to obtain a controlled deviation $z$. This operation introduces directional variation while keeping the perturbation magnitude bounded, resulting in a new trajectory that remains close to but deviates from the correct path. The perturbed trajectory is then rescaled to match the magnitude of the original trajectory and combined with it to form a misleading negative feature:
\begin{equation}
z = (\cos\theta_{\text{neg}} \cdot \Delta_{\text{norm}} + \sin\theta_{\text{neg}} \cdot \eta_{\text{norm}}) \cdot ||\Delta||,
\end{equation}
\begin{equation}
S_{\text{neg}} = S_{I_{\text{hint}}} + z = S_{I} + \Delta + z.
\end{equation}

This procedure produces a set of positive and negative samples distributed around the true reasoning trajectory. During training, we repeat the above process $N_{\text{neg}}$ times and optimize the latent representations using the InfoNCE loss~\cite{he2020momentum}, which encourages closeness to positive samples and separation from the structured negative samples:
\begin{equation}
\mathcal{L}_{\text{contras}} = -\frac{1}{K} \sum_i^K \log \left( \frac{e^{\text{sim}(h_i, S_{I_{\text{hint}}})/\tau}}{e^{\text{sim}(h_i, S_{I_{\text{hint}}})/\tau} + \sum_{j}^{N_{\text{neg}}} e^{\text{sim}(h_i, S_{\text{neg}}^j)/\tau}} \right),
\end{equation}
where $\tau$ represents the temperature. In addition, we introduce a cross-entropy loss on the answer to preserve the model's text generation capability:
\begin{equation}
\mathcal{L}_{\text{contrasSFT}} = \lambda_2 \mathcal{L}_{\text{contras}} + \mathcal{L}_{\text{CE}}.
\end{equation}

\subsection{Latent-Trajectory Contrastive Reward}
Finally, we further improve the reinforcement learning stage to alleviate the homogenization problem in the latent space. Unlike existing latent reasoning methods that mainly rely on format rewards and final answer rewards at the token distribution level, our method introduces latent-trajectory optimization to encourage more diverse and exploratory latent representations.

Specifically, the importance sampling term in the GRPO (Group Relative Policy Optimization) objective $r_t(\theta) = \frac{\pi_\theta(a_t \mid s_t)}{\pi_{\theta_{\text{old}}}(a_t \mid s_t)}$
indicates that the optimization aims to increase the generation probability of tokens corresponding to correct answers, thereby encouraging the model to produce higher-quality outputs in subsequent reasoning steps. Inspired by this observation, the optimization of hidden states should also move closer to the latent representations of correct samples. However, enforcing hard alignment with the hidden states corresponding to correct answers may suppress the exploration ability of the latent space. Therefore, it is necessary to introduce a contrastive learning mechanism that pulls positive samples closer while pushing negative samples apart, enabling the learning of more flexible latent representations.

Based on this motivation, this paper designs a latent-trajectory contrastive reward that provides relative preference signals among latent trajectories rather than enforcing predefined trajectory targets (as shown in Figure~\ref{fig:framework}(d)). Specifically, the cosine similarity $\text{sim}$ between latent states is first defined as
\begin{equation}
Tr_i = { h_i^1, h_i^2, \dots, h_i^K },
\end{equation}
\begin{equation}
\overline{\text{sim}}(Tr_i, Tr_j) = \frac{1}{K}\sum_t \overline{\text{sim}}(h_t^i, h_t^j), \quad \text{where } \overline{\text{sim}} = \frac{1 + \text{sim}}{2},
\end{equation}
where the similarity is obtained by averaging the similarities of latent tokens at corresponding positions, and is normalized to the interval $[0, 1]$ via the transformation $\overline{\text{sim}} = (1 + \text{sim}) / 2$ to ensure stable reward scaling. During the rollout phase, the model collects latent trajectories corresponding to both correct and incorrect samples generated under the current policy, and performs an additional forward pass to obtain latent representations $Tr_{\text{now}}$ under consistent conditions. When correct answers are present, one latent trajectory is selected as the positive sample, while the remaining incorrect latent trajectories are treated as negative samples, and an InfoNCE-based reward is constructed at the hidden state level. When no correct answer exists, only a repulsion constraint from negative samples is applied. When no incorrect answer exists, only alignment with one latent trajectory is performed. The reward is defined as
\begin{equation}
    R_{\text{latent}} = 
    \begin{cases}
    \frac{ e^{\overline{\text{sim}}(Tr_{\text{now}}, Tr_{\text{pos}})/\tau} }{ e^{\overline{\text{sim}}(Tr_{\text{now}}, Tr_{\text{pos}})/\tau} + \sum_{j}^{n_{\text{neg}}} e^{\overline{\text{sim}}(Tr_{\text{now}}, Tr_j)/\tau} }, & \small\text{if pos \& neg mixed,} \\
    \frac{ 1 }{ 1 + \sum_{j}^{n_{\text{neg}}} e^{\overline{\text{sim}}(Tr_{\text{now}}, Tr_j)/\tau} }, & \small\text{if only neg,} \\
    \overline{\text{sim}}(Tr_{\text{now}}, Tr_{\text{pos}})/\tau, & \small\text{if only pos.}
\end{cases}
\end{equation}

Through this case-wise design, the proposed reward balances exploration and discrimination in the latent space during reinforcement learning. Finally, the proposed latent-trajectory contrastive reward is jointly incorporated into the GRPO training framework together with the original format $R_{\text{format}}$ and textual rewards $R_{\text{correct}}$. The format reward combines a token count ratio term for \texttt{<|latent\_pad|>} scaled by 0.5 with a \texttt{"\textbackslash boxed\{\}"} format match bonus of 0.5. The textual reward is binary: 1 if the final answer matches the ground truth, 0 otherwise. This joint reward preserves generation quality while enabling effective optimization of the latent reasoning process.
\begin{equation}
    R_{\text{total}}=R_{\text{format}}+R_{\text{correct}}+R_{\text{latent}}.
\end{equation}
\section{Experiments}
\subsection{Experiment Settings}
\textbf{Training Datasets and Benchmarks.} This paper evaluates the model on diverse visual reasoning tasks. For VSP (Visual Spatial Planning), the goal is to measure path planning and spatial reasoning in maze environments. We use Mirage~\cite{yang2025machine} VSP training data and benchmarks for base evaluation, and further introduce level 7 and level 8 datasets built on DiffThinker~\cite{he2025diffthinkergenerativemultimodalreasoning} to test generalization under higher complexity. For visual puzzle tasks, Jigsaw evaluates the ability to model global semantics and recover missing regions from incomplete input, while Tertis examines constrained spatial completion through a Tetris-like mechanism. For these tasks, we use two Zebra-CoT~\cite{li2026zebracot} datasets split into training and test sets for systematic analysis. To further assess cross-task generalization and \ours's exploratory advantage in latent visual reasoning, we also include multiple visual reasoning benchmarks, including VisPuzzle (from Thinkmorph~\cite{gu2026thinkmorph}), MMVP~\cite{tong2024eyeswideshutexploring}, MMStar~\cite{chen2024rightwayevaluatinglarge}, V*~\cite{wu2023vguidedvisualsearch} and CV-Bench~\cite{tong2024cambrian1fullyopenvisioncentric}, enabling comprehensive evaluation.

\textbf{Baselines.} We use Qwen2.5-VL-7B-Instruct~\cite{bai2025qwen25vltechnicalreport} (denoted as Qwen2.5-VL-7B) as the base model for both supervised fine-tuning and reinforcement learning. For evaluation, closed-source baselines include GPT-4o~\cite{openai2024gpt4ocard} and Gemini-2.5-flash~\cite{comanici2025gemini25pushingfrontier}, while the open-source comparison is InternVL3.5-8B-Instruct~\cite{wang2025internvl35advancingopensourcemultimodal} (denoted as InternVL3.5-8B) . In addition, Qwen2.5-VL-7B is further trained on the same dataset with both supervised fine-tuning and GRPO~\cite{shao2024deepseekmathpushinglimitsmathematical}, and the Mirage latent training paradigm is implemented under the same data setting for direct comparison, enabling a systematic assessment of the proposed method. To further examine generalization, LVR~\cite{li2026latent} is introduced as a complementary baseline, which achieves region-level alignment through ROI indexing.

\textbf{Implementation Details.} Training is conducted on four NVIDIA H200 GPUs with a global batch size of 4. To ensure reproducibility, the random seed is fixed at 42. For the VSP task, the learning rates across different training stages are set to 5e-5, 5e-5, and 5e-6, respectively, with a linear decay schedule. The number of latent tokens is consistently set to 8 and remains fixed during inference to maintain alignment between training and inference. During the SFT stage, $\theta$ is sampled from [$\frac{\pi}{2}$, $\pi$], and the number of negative samples $N_{\text{neg}}$ is set to 8. In addition, both supervised fine-tuning and reinforcement learning are implemented using the TRL framework.

\subsection{Main Results}
\begin{table}[t]
  \caption{\textbf{Main results across multiple visual task benchmarks}, comparing \ours{} with other VLMs and rigidly aligned latent baseline Mirage. Results are reported in accuracy (\%), and best results are highlighted in \textbf{bold}.}
  \label{main_table}
  \centering
  \resizebox{1.0\textwidth}{!}{
      \begin{tabular}{lccccccccccc}
        \toprule
        \multirow{3}{*}{\textbf{Model}} & \multicolumn{9}{c}{\textbf{VSP}} & \multirow{3}{*}{\textbf{Jigsaw}} & \multirow{3}{*}{\textbf{Tertis}} \\
        \cmidrule(r){2-10}
         & \multicolumn{5}{c}{Seen} & \multicolumn{3}{c}{Unseen} & \multirow{2}{*}{Total}\\
         \cmidrule(r){2-6}
         \cmidrule(r){7-9}
         & Level 3 & Level 4 & Level 5 & Level 6 & Avg. & Level 7 & Level 8 & Avg. & & & \\
        \midrule
        \rowcolor{gray!15}
        \multicolumn{12}{c}{\textit{\textbf{Proprietary Model}}} \\
        GPT-4o~\cite{openai2024gpt4ocard} & 67.00 & 46.00 & 24.00 & 17.00 & 38.50 & 11.00 & 6.00 & 8.50 & 28.50 & 36.00 & 30.67 \\
        Gemini-2.5-flash~\cite{comanici2025gemini25pushingfrontier} & 95.00 & 86.00 & 62.00 & 30.00 & 62.00 & 32.00 & 18.00 & 25.00 & 53.83 & 43.33 & 31.33 \\
        
        \rowcolor{gray!15}
        \multicolumn{12}{c}{\textit{\textbf{Open source model}}} \\
        InternVL3.5-8B~\cite{wang2025internvl35advancingopensourcemultimodal} & 20.00 & 13.00 & 3.00 & 2.00 & 9.50 & 2.00 & 0.00 & 1.00 & 6.67 & 19.33 & 22.00 \\
        Qwen2.5-VL-7B~\cite{bai2025qwen25vltechnicalreport} & 11.00 & 7.00 & 7.00 & 1.00 & 6.50 & 1.00 & 0.00 & 0.50 & 4.50 & 24.00 & 24.00 \\
        \quad w/ CoT & 20.00 & 8.00 & 5.00 & 3.00 & 9.00 & 3.00 & 2.00 & 2.50 & 6.83 & 24.67 & 21.33 \\
        \quad w/ vanilla SFT & 91.00 & 85.00 & 71.00 & 50.00 & 74.30 & 36.00 & 29.00 & 29.00 & 59.17 & 67.33 & 42.67  \\
        \quad w/ vanilla SFT + GRPO & 89.00 & 80.00 & 66.00 & 49.00 & 71.00 & 34.00 & 14.00 & 24.00 & 55.33 & 68.00 & 41.33  \\
        
        \rowcolor{gray!15}
        \multicolumn{12}{c}{\textit{\textbf{Latent model}}} \\
        Mirage~\cite{yang2025machine} & 90.00 & 84.00& 67.00 & 55.00 & 74.00 & 39.00 & 25.00 & 32.00 & 60.00 & 70.00 & 43.33  \\
        \rowcolor{OursBlue}
        \textbf{\ours (Ours)} & \textbf{94.00} & \textbf{92.00} & \textbf{80.00} & \textbf{57.00} & \textbf{80.70} & \textbf{43.00} & \textbf{29.00} & \textbf{36.00} & \textbf{65.83} & \textbf{78.00} & \textbf{44.67} \\
        $\Delta$ vs Mirage & \textcolor{DeltaGreen}{+4.00} & \textcolor{DeltaGreen}{+8.00} & \textcolor{DeltaGreen}{+13.00} & \textcolor{DeltaGreen}{+2.00} & \textcolor{DeltaGreen}{+6.70} & \textcolor{DeltaGreen}{+4.00} & \textcolor{DeltaGreen}{+4.00} & \textcolor{DeltaGreen}{+4.00}  & \textcolor{DeltaGreen}{+5.83} & \textcolor{DeltaGreen}{+8.00} & \textcolor{DeltaGreen}{+1.34}\\
        \bottomrule
      \end{tabular}
  }
\end{table}
From the experimental results in Table~\ref{main_table}, we observe that the proposed \ours{} framework consistently outperforms all competing baselines across multiple visual reasoning benchmarks, demonstrating stable and significant performance gains. Specifically, compared with both the vanilla SFT approach and the reinforcement learning based GRPO method, \ours{} achieves superior results on all evaluation metrics, indicating that latent space contrastive optimization effectively enhances the model’s visual reasoning capability. Furthermore, when compared with Mirage, which relies on a hard alignment mechanism, \ours{} consistently surpasses it across all task settings, with overall improvements ranging from 1.34\% to 8.00\%. Notably, on the Jigsaw task, which requires stronger implicit visual imagination, the advantage of \ours{} becomes more pronounced, achieving an 8.00\% improvement over Mirage. These results suggest that \ours{} introduces a more discriminative latent feature learning mechanism, which alleviates the representational constraints introduced by hard alignment in existing latent space models, thereby facilitating the learning of more exploratory latent tokens and enabling better modeling and reasoning of visual semantics.

\subsection{Generalization on out of domain visual benchmarks}
To further analyze the role of the exploratory latent contrastive objective in latent reasoning, we evaluate the \ours{} model on multiple out-of-domain benchmarks and compare it with two latent space models that adopt hard alignment constraints on visual features, namely Mirage and LVR. The results show that, \ours{} significantly improves the generalization ability of latent reasoning models and consistently achieves the best performance across all evaluated datasets. Specifically, as shown in Table~\ref{main_table}, on the VSP benchmark, the \ours{} model trained on the VSP task demonstrates strong robustness under the more challenging unseen settings, achieving an average score of 36.00\%, which improves over Mirage by 4.00\%. This indicates stronger transferability under distribution shift. As shown in Table~\ref{ood_table}, we evaluate the \ours{} model trained on the Jigsaw task. On the V* benchmark, \ours{} improves over Mirage by 4.71\%, reaching an accuracy of 80.63\%, and slightly outperforms LVR (8 steps), which achieves 80.10\%. Moreover, on MMVP and MMStar, which require fine-grained visual perception and multimodal understanding, \ours{} achieves the highest scores of 61.47\% and 72.00\%, respectively.

These results suggest that latent tokens in \ours{} exhibit stronger flexibility in adapting to unseen visual scenarios, whereas Mirage and LVR, which rely on predefined feature spaces, show limited adaptability under distribution shift. This further indicates that the latent contrastive objective helps prevent the latent space from collapsing into a homogeneous semantic manifold. By encouraging more exploratory trajectories in the hidden state space, \ours{} effectively transfers the reasoning patterns learned from the training task to complex real-world visual understanding benchmarks.

\begin{table}[t]
  \caption{\textbf{Out-of-domain visual understanding benchmarks comparing \ours{} with other latent reasoning models}. LVR uses official checkpoints, while Mirage is trained on the same data as our method. Results are reported in accuracy (\%), with the best performance highlighted in \textbf{bold}.}
  \label{ood_table}
  \centering
  \resizebox{0.60\textwidth}{!}{
      \begin{tabular}{lccccc}
        \toprule
        \textbf{Model} & {\textbf{VisPuzzle}} & \textbf{V*} & \textbf{MMVP} & \textbf{MMStar} & \textbf{CV-Bench}\\
        \midrule
        \rowcolor{OursBlue}
        \ours  & \textbf{37.00} & \textbf{80.63} & \textbf{72.00} & \textbf{61.47} & \textbf{76.95} \\
        Mirage & 36.75 & 75.92 & 68.33 & 55.27 & 72.40 \\
        \quad $\Delta$ & 
        \textcolor{DeltaGreen}{+0.25} & \textcolor{DeltaGreen}{+4.71} & \textcolor{DeltaGreen}{+3.67} & \textcolor{DeltaGreen}{+6.20} & 
        \textcolor{DeltaGreen}{+4.55} \\
        LVR (8 steps) & \textbf{37.00} & 80.10 & 69.33 &  58.07 & 75.51 \\
        \quad $\Delta$ & 
        \textcolor{gray!50}{+0.00} & \textcolor{DeltaGreen}{+0.53} & \textcolor{DeltaGreen}{+2.67} & \textcolor{DeltaGreen}{+3.40} &
        \textcolor{DeltaGreen}{+1.44} \\
        \bottomrule
      \end{tabular}
    }
\end{table}

\subsection{Ablation Study}

\textbf{The latent contrastive framework improves visual reasoning across both seen and unseen settings.}
Table~\ref{ablation} evaluates the contributions of the key components in \ours{}.
Compared with the warm-up-only variant, which removes both contrastive SFT and trajectory-reward GRPO, the full framework improves VSP Seen and Unseen accuracy by 5.20 and 6.50 percentage points, respectively, and raises the overall VSP score from 60.17\% to 65.83\%.
It also improves Jigsaw and Tetris accuracy by 3.30 and 4.67 points, respectively.
These consistent gains support the effectiveness of the overall training framework, with the larger improvement on unseen VSP tasks suggesting better generalization beyond the training settings.

\begin{table}[t]
  \caption{\textbf{Ablation study of key design components}. \textit{w/o Traj Reward GRPO} removes the final GRPO stage with latent trajectory rewards; \textit{+ w/o Contras SFT} retains only the warm-up training stage; \textit{w/ Traj GRPO Only} directly applies GRPO with latent trajectory rewards after warm-up; \textit{w/ Normal GRPO} applies GRPO without latent trajectory rewards.}
  \label{ablation}
  \centering
  \resizebox{1.0\textwidth}{!}{
      \begin{tabular}{lccccccccccc}
        \toprule
        \multirow{3}{*}{\textbf{Model}} & \multicolumn{9}{c}{\textbf{VSP}} & \multirow{3}{*}{\textbf{Jigsaw}} & \multirow{3}{*}{\textbf{Tertis}} \\
        \cmidrule(r){2-10}
         & \multicolumn{5}{c}{Seen} & \multicolumn{3}{c}{Unseen} & \multirow{2}{*}{Total}\\
         \cmidrule(r){2-6}
         \cmidrule(r){7-9}
         & Level 3 & Level 4 & Level 5 & Level 6 & Avg. & Level 7 & Level 8 & Avg. & & & \\
        \midrule
        \rowcolor{OursBlue}
        \textbf{\ours (Ours)} & 94.00 & \textbf{92.00} & \textbf{80.00} & \textbf{57.00} & \textbf{80.70} & \textbf{43.00} & \textbf{29.00} & \textbf{36.00} & \textbf{65.83} & \textbf{78.00} & \textbf{44.67} \\
        \quad w/o Traj Reward GRPO & \textbf{95.00} & 91.00 & 75.00 & 54.00 & 78.70 & 39.00 & 27.00 & 33.00 & 63.50 & 78.00 & 42.67 \\
        \qquad + w/o Contras SFT & 92.00 & 87.00 & 72.00 & 51.00 & 75.50 & 35.00 & 24.00 & 29.50 & 60.17 & 74.70 & 40.00 \\
        \quad w/ Traj GRPO Only & 92.00 & 87.00 & 75.00 & 44.00 & 74.50 & 36.00 & 27.00 & 31.50 & 60.17 & 75.33 & 40.00 \\
        \quad w/ Normal GRPO & 92.00 & 87.00 & 69.00 & 48.00 & 74.00 & 39.00 & 27.00 & 33.00 & 60.33 & 76.00 & 38.67 \\
        \bottomrule
      \end{tabular}
  }
\end{table}


\textbf{Contrastive SFT and trajectory-reward GRPO make complementary contributions.}
To examine the role of Stage 2 before GRPO optimization, we directly apply trajectory-reward GRPO after warm-up (\textit{w/ Traj GRPO Only}).
Compared with this variant in Table~\ref{ablation}, the full framework improves VSP, Jigsaw, and Tertis scores by 5.66, 2.67, and 4.67 points, respectively, supporting the importance of learning task-relevant latent representations through contrastive SFT before subsequent trajectory-level optimization.
Stage 3 provides a further benefit: compared with the variant without trajectory-reward GRPO (\textit{w/o Traj Reward GRPO}), the full model achieves a 2.33 percentage-point improvement in the overall VSP score.
By contrast, applying standard GRPO without the latent trajectory reward (\textit{w/ Normal GRPO}) reduces this score to 60.33\%.
Together, these results indicate that the gains depend on both the preceding contrastive SFT stage and the trajectory-specific reward, supporting a training design that combines representation-level contrastive learning with trajectory-level optimization.


\begin{table}[t]
  \centering

  \begin{minipage}[t]{0.58\textwidth}
    \centering
    \captionof{table}{Comparison of different negative construction strategies for the latent contrastive objective.}
    \label{tab:neg_abla}
    \resizebox{\linewidth}{!}{
      \begin{tabular}{lcccc}
        \toprule
        \textbf{Neg. construction} & \textbf{VSP Seen} & \textbf{VSP Unseen} & \textbf{VSP Total} & \textbf{Jigsaw} \\
        \midrule
        Random batch-pool  & 74.00 & 27.00 & 58.33 & 76.67 \\
        Incorrect-image-based & 77.50 & 30.00 & 61.67 & 78.00 \\
        Noise-based & 75.20 & 30.50 & 60.33 & 75.33 \\
        \rowcolor{OursBlue}
        Angle-based & \textbf{78.70} & \textbf{33.00} & \textbf{63.50} & \textbf{78.00} \\
        \bottomrule
      \end{tabular}
    }
  \end{minipage}
  \hfill
  \begin{minipage}[t]{0.37\textwidth}
    \captionof{table}{Semantic association of constructed negatives.}
    \label{tab:homo}
    \centering
    \resizebox{\linewidth}{!}{
      \begin{tabular}{lccc}
        \toprule
        \textbf{Measure} & \textbf{VSP} & \textbf{Jigsaw} & \textbf{Tertis} \\
        \midrule
        Neg. Hint Cos. & 0.97 & 0.89 & 0.91 \\
        RIA & 0.57 & 0.23 & 0.17 \\
        \bottomrule
      \end{tabular}
    }
  \end{minipage}
  \label{tab:ablation}
\end{table}

\begin{figure}[t]
    \centering
    \subfigure[Negative construction sensitivity.]{
        \includegraphics[width=0.56\linewidth]{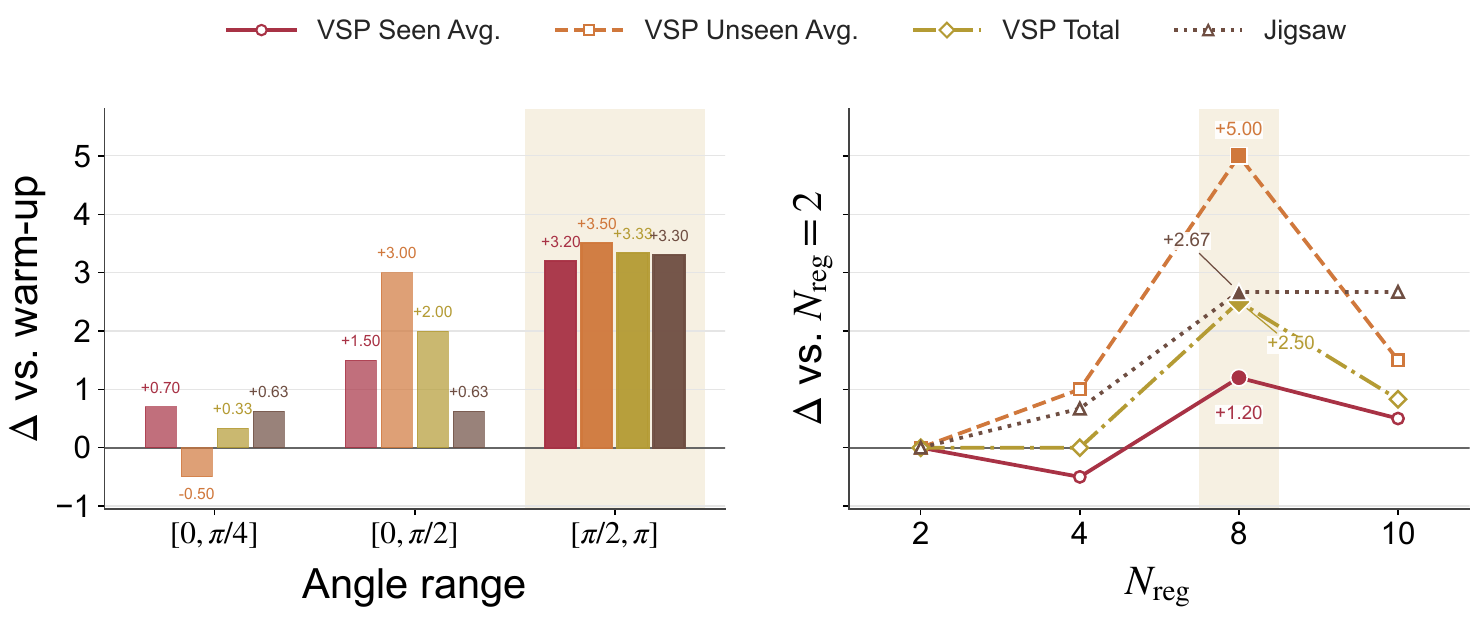}
        \label{fig:negative_sensitivity}
    }
    \hfill
    \subfigure[Latent representation uniformity.]{
        \includegraphics[width=0.4\linewidth]{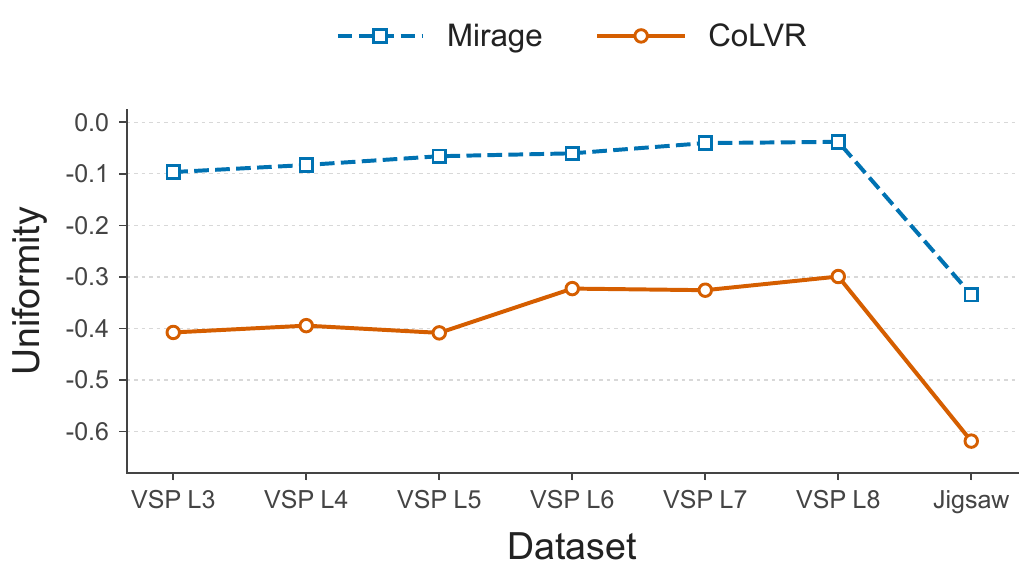}
        \label{fig:latent_uniformity}
    }

    \caption{\textbf{Sensitivity of negative construction and analysis of latent-space homogenization.}
        (a) Effects of the angle range (left) and the number of negatives
        (right), measured as accuracy changes in percentage points relative
        to warm-up and $N_{\mathrm{neg}}=2$, respectively.
        Shaded regions indicate the selected settings.
        (b) Hyperspherical uniformity of normalized latent hidden states
        across VSP levels and Jigsaw; lower values indicate less concentrated
        representations. \ours{} consistently achieves lower scores than Mirage.}
    
    \label{fig:sensitivity_and_uniformity}
\end{figure}

\textbf{Angle-based perturbation provides effective negative supervision for contrastive SFT.}
Table~\ref{tab:neg_abla} compares angle-based perturbations with batch-pool negatives from other samples, incorrect-hint image features, and Gaussian-noise perturbations.
Our strategy achieves the highest VSP Total accuracy of 63.50\%, exceeding these alternatives by 5.17, 1.83, and 3.17 percentage points, respectively, without requiring task-specific incorrect images during training.
To assess the semantic relevance of the constructed negatives, we use incorrect solution-hint images generated with Gymnasium for VSP and provided by Zebra-CoT for the other tasks as references.
Table~\ref{tab:homo} shows high cosine similarity between constructed negatives and incorrect-hint features.
To account for their shared correct-hint component, we define Residual Incorrect Association (RIA) as the cosine similarity between their residuals after projecting out the correct-hint direction, following the principle of first-order partial correlation~\cite{baba2004partial}.
Positive RIA values across all three tasks indicate that the association persists beyond this shared component, supporting the semantic relevance of the constructed negatives.

\textbf{Sensitivity to the angle range and number of negatives.}
Figure~\ref{fig:negative_sensitivity} (left) shows that $[\frac{\pi}{2},\pi]$ yields the largest gains over warm-up across all evaluated metrics.
With the noise component orthogonal to the correct trajectory direction $\Delta$, this range ensures a non-positive projection onto $\Delta$, producing orthogonal or backward deviations.
For the number of negatives $N_{\mathrm{\text{neg}}}$ (right), increasing from two to eight improves all metrics, particularly unseen VSP, whereas increasing to ten reduces VSP performance without improving Jigsaw.
These results support $\theta\in[\frac{\pi}{2},\pi]$ and $N_{\mathrm{neg}}=8$ as our default configuration, which performs best overall among the tested settings.

\subsection{Exploratory Capability Analysis}

\begin{figure}[t]
    \centering

    \subfigure[UMAP Visualization of Latent Tokens in Jigsaw task.]{
        \includegraphics[width=0.54\linewidth]{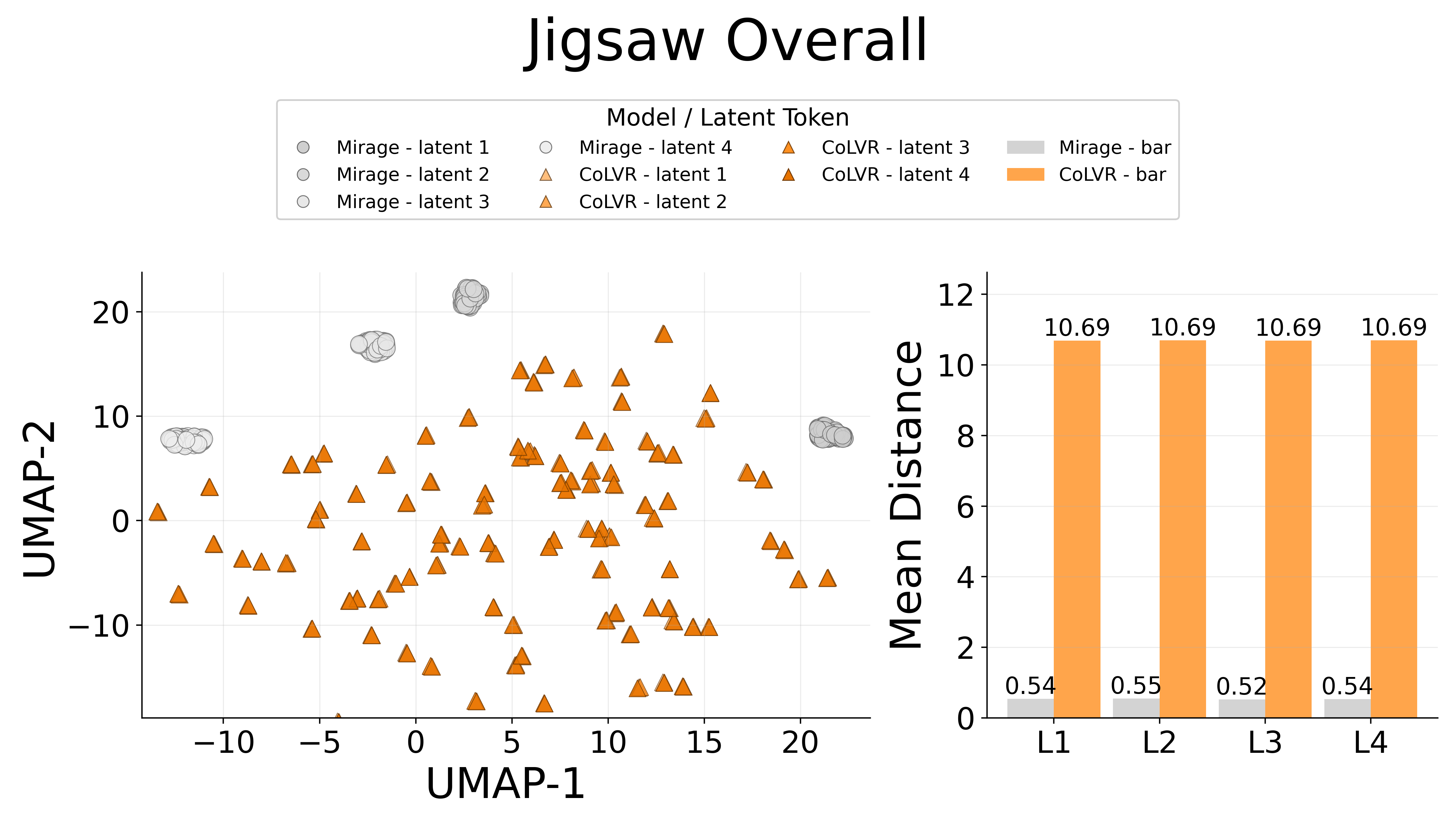}
        \label{fig:umap_jigsaw_compare}
    }
    \hfill
    \subfigure[Inference Noise Injection.]{
        \includegraphics[width=0.42\linewidth]{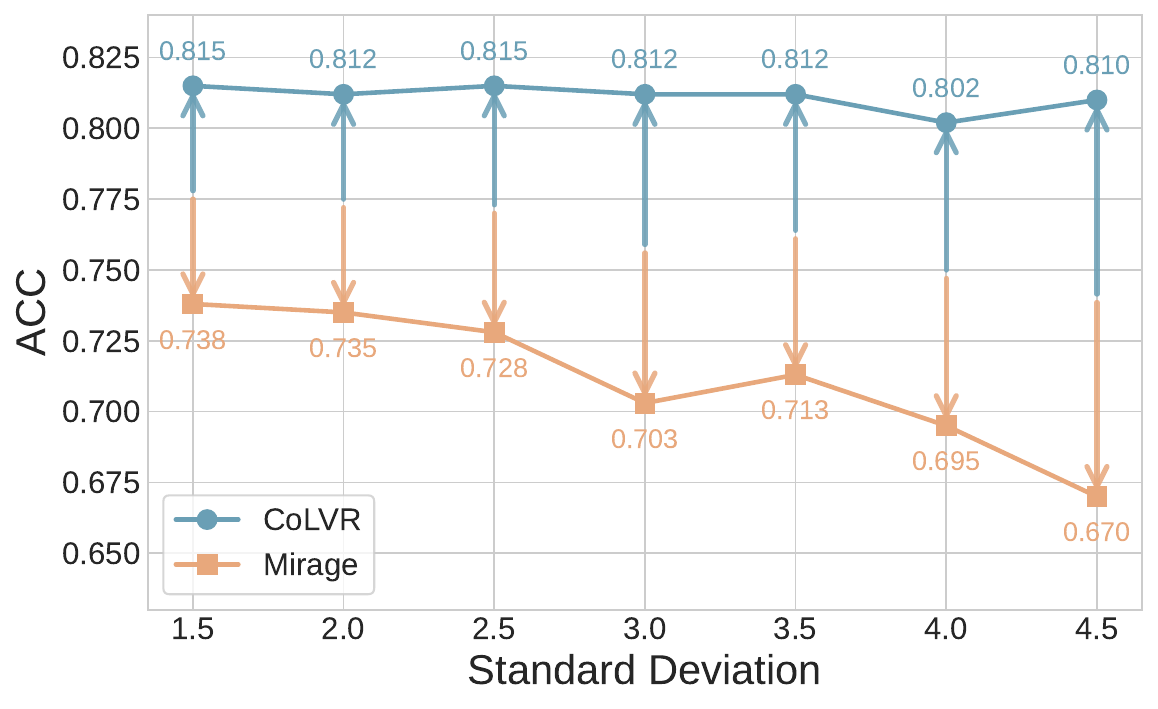}
        \label{fig:colvr_vs_mirage}
    }

    \caption{(a) \textbf{UMAP Visualization of Latent Tokens in Jigsaw task}, Mirage and CoLVR exhibit the same trend as observed on VSP. (b) \textbf{Inference Noise Injection}. We introduce random noise $\epsilon \sim \mathcal{N}(0, \sigma^2)$ into the latent tokens during the inference phase to evaluate their sensitivity and explorative capabilities.}
    
    \label{fig:noise_std}
\end{figure}

\textbf{\ours{} learns a substantially more exploratory and discriminative latent space than rigid alignment-based methods}. As shown in Figure~\ref{fig:caption_b} and ~\ref{fig:umap_jigsaw_compare}, we compare the UMAP visualizations of the first four latent tokens across different VSP levels and Jigsaw for Mirage and \ours{}. The visualizations show more dispersed representations for \ours{},
with average distances to the corresponding cluster centroids of
approximately 10.5--12.7, compared with 0.5--0.7 for Mirage.

To quantify concentration in the original latent space, we adopt
hyperspherical uniformity~\cite{wang2020understandinguniformity}:
$\mathcal{L}_{\mathrm{\text{uniform}}}
= \log \mathbb{E}_{i \neq j}
[\exp(-t\lVert z_i-z_j\rVert_2^2)]$,
where $z_i$ is an $L_2$-normalized latent hidden state and $t=2$.
Lower values indicate more uniformly distributed representations.
As shown in Figure~\ref{fig:latent_uniformity}, \ours{} achieves
consistently lower scores across all evaluated VSP levels and Jigsaw,
supporting reduced latent-space homogenization and aligning with
our objective of enabling more exploratory latent reasoning.
 
\textbf{Contrastive latent optimization enables \ours{} to maintain robust reasoning under broader latent space exploration}. To further evaluate the exploration capability learned by \ours{} in the latent space, we inject Gaussian noise $\mathcal{N}(0, \sigma^2)$ into the propagated last hidden state during latent reasoning, and assess the model performance on VSP under controlled perturbations ( Figure~\ref{fig:colvr_vs_mirage}). Such noise forces the model to explore a broader region of the latent space. Under such perturbations, \ours{} maintains stable performance, whereas Mirage exhibits severe degradation. These results suggest that contrastive learning over positive and negative samples enables \ours{} to learn more exploratory and robust latent representations, sustaining effective reasoning under wider latent space exploration. In contrast, methods based on hard alignment objectives rely on highly compressed latent spaces, making them substantially more sensitive to perturbations.

\section{Conclusion}
\label{section:conclusion}
In this work, we address the issue of latent space collapse in multimodal reasoning by examining how rigid hard-alignment objectives can limit the exploratory behavior of latent tokens. To mitigate this problem, we propose \ours{}, a framework that moves from static feature matching toward contrastive trajectory optimization. By introducing angle-based perturbations during supervised fine-tuning and incorporating latent trajectory contrastive reward, \ours{} helps improve the latent hidden state exploratory capability. Our experiments across visual reasoning and out-of-domain benchmarks, together with UMAP visualizations, hyperspherical uniformity measurements and noise-injection tests, consistently show that \ours{} leads to a more exploratory and flexible latent space. Overall, this approach provides a scalable way to improve the flexibility and generalization of latent reasoning, moving beyond hard alignment constraints toward more exploratory multimodal reasoning.


\textbf{Limitations}. Despite its effectiveness, \ours{} relies on result hints for positive construction, increasing supervision costs.
Future work could bootstrap contrastive SFT with correctness-filtered pseudo-hints from model rollouts.
Limited semantic grounding of latent tokens also hinders interpretability and diagnostic analysis, motivating lightweight concept-level probing.
\begin{ack}
This work is supported by Shanghai Artificial Intelligence Laboratory and Smart Grid-National Science and Technology Major Project (No. 2025ZD0803600, 2025ZD0803603). We also thank Zitang Gou from Fudan University for his support.
\end{ack}

\newpage
\small
\bibliographystyle{plain}
\bibliography{ref}

\newpage
\appendix
\section{Appendix}
\subsection{Detailed Experimental Settings}
All training procedures for CoLVR are conducted using the Accelerate library to enable efficient multi-GPU parallel training. Across all stages, we uniformly set the warm-up steps to 10, gradient accumulation steps to 1, and weight decay to 0.01. The AdamW optimizer is adopted throughout, with training precision set to bf16 and gradient checkpointing enabled to reduce memory consumption. For a fair comparison, Mirage is trained under identical settings to CoLVR. The learning rate and number of training epochs for each stage in CoLVR are listed in the Table~\ref{experiment_setup_colvr} below:
\begin{table}[h]
  \caption{\textbf{The detailed experiment settings of CoLVR training process.}}
  \label{experiment_setup_colvr}
  \centering
  \resizebox{1.0\textwidth}{!}{
      \begin{tabular}{lccccccccc}
        \toprule
        \multirow{2}{*}{Stage} & \multicolumn{3}{c}{VSP} & \multicolumn{3}{c}{Jigsaw} & \multicolumn{3}{c}{Tertis} \\
        \cmidrule(r){2-4}
        \cmidrule(r){5-7}
        \cmidrule(r){7-9}
         & learning rate & epochs & batch size & learning rate & epochs & batch size & learning rate & epochs & batch size \\
        \midrule
        Warm up  & 5e-5 & 10 & 4 & 2e-5 & 8 & 4 & 1e-5 & 8 & 4 \\
        Latent Contrastive SFT & 5e-5 & 10 & 4 & 2e-5 & 8 & 4 & 1e-5 & 8 & 4 \\
        Latent Trajectory Contrastive GRPO & 5e-6 & 5 & 4 & 2e-6 & 5 & 4 & 2e-6 & 5 & 4\\

        \bottomrule
      \end{tabular}
  }
\end{table}

During the warm-up stage, the weight of the cross-entropy (CE) loss is set to $\lambda_1 = 0.3$.
In the latent contrastive SFT stage, the weight of the contrastive loss is set to $\lambda_2 = 2$.
The remaining hyperparameters used in the GRPO stage are summarized in Table~\ref{rl_hyper}.
\begin{table}[h]
  \caption{\textbf{Hyperparameters for RL.}}
  \label{rl_hyper}
  \centering
  \resizebox{0.8\textwidth}{!}{
      \begin{tabular}{cccccc}
        \toprule
        Rollout size & Temperature & Max prompt length & Max completion length & $\beta$ & Latent trajectory temperture $\tau$ \\
        \midrule
        4  & 1.2 & 2048 & 1024 & 0.04 & 0.5 \\
        \bottomrule
      \end{tabular}
  }
\end{table}

Meanwhile, during training, we froze the parameters of the visual encoder and trained all remaining parameters of the MLLM, ensuring that the pretrained visual representations remain intact while the model adapts to the target tasks. Regarding the data, the amounts of data used for training are shown in Table~\ref{dataset_stac}. For RL training, the VSP data are generated using DiffThinker code, while the Jigsaw and Tetris data are randomly sampled from Zebra-CoT. We ensure that the RL training data do not overlap with either the SFT training dataset or the test dataset.Notably, since the images provided in zebra-jigsaw have relatively large resolutions, we resized them proportionally to a maximum size of 1024 pixels to ensure training stability and prevent out-of-memory issues. 
\begin{table}[h]
  \caption{\textbf{Dataset Statistics.}}
  \label{dataset_stac}
  \centering
  \resizebox{0.4\textwidth}{!}{
      \begin{tabular}{lccc}
        \toprule
        Task & Warm-up/SFT & RL & Test \\
        \midrule
        VSP  & 1000 & 500 & 400 \\
        VSP (unseen)  & / & / & 200 \\
        Jigsaw & 1516 & 500 & 150 \\
        Tertis & 1500 & 500 & 150 \\

        \bottomrule
      \end{tabular}
  }
\end{table}

Mirage was trained on the same training data as CoLVR to ensure a fair comparison. During testing, we retained Mirage's original inference protocol: the upper limit of latent tokens was set to 8, and the model was allowed to freely determine how many latent tokens to output based on its internal halting mechanism. In contrast, CoLVR was fixed to output exactly 8 latent tokens, providing a controlled baseline for evaluating the effect of adaptive token allocation.

\subsection{Dynamic Latent Inference Comparison}
Given the slight difference in inference modes between CoLVR and Mirage, we adapted CoLVR's inference protocol to match that of Mirage and re-evaluated it on the VSP, Jigsaw, and Tetris tasks. The results are presented in Table~\ref{dynamic_inference} below. It can be observed that dynamically determining versus fixing the number of latent tokens has no measurable impact on CoLVR's performance, indicating that CoLVR learns latent representations that are robust to token allocation and do not rely on a precise token count to maintain reasoning quality.
\begin{table}[ht]
  \caption{\textbf{Main results across multiple visual task benchmarks on the same inference mode.}}
  \label{dynamic_inference}
  \centering
  \resizebox{\textwidth}{!}{
      \begin{tabular}{lccccccccccc}
        \toprule
        \multirow{3}{*}{Model} & \multicolumn{9}{c}{VSP} & \multirow{3}{*}{Jigsaw} & \multirow{3}{*}{Tertis} \\
        \cmidrule(r){2-10}
         & \multicolumn{5}{c}{Seen} & \multicolumn{3}{c}{Unseen} & \multirow{2}{*}{Total}\\
         \cmidrule(r){2-6}
         \cmidrule(r){7-9}
         & Level 3 & Level 4 & Level 5 & Level 6 & Avg. & Level 7 & Level 8 & Avg. & & & \\
        \midrule
        \ours & \textbf{94.00} & \textbf{92.00} & \textbf{80.00} & \textbf{57.00} & \textbf{80.70} & \textbf{43.00} & \textbf{29.00} & \textbf{36.00} & \textbf{65.83} & \textbf{78.00} & \textbf{44.67} \\
        \ours{} (dynamic) & \textbf{94.00} & \textbf{92.00} & \textbf{80.00} & \textbf{57.00} & \textbf{80.70} & \textbf{43.00} & \textbf{29.00} & \textbf{36.00} & \textbf{65.83} & \textbf{78.00} & \textbf{44.67} \\
        Mirage & 93.00 & 83.00& 76.00 & 51.00 & 76.00 & 39.00 & 25.00 & 32.00 & 61.17 & 70.00 & 43.33 \\
        \bottomrule
      \end{tabular}
  }
\end{table}


To visualize how latent tokens explore an image, we compare the attention maps of the first four latent tokens in Mirage and CoLVR across the four levels of VSP. These maps, shown as heatmaps in Figure~\ref{fig:appendix_attenmap}, reveal a stark contrast in spatial coverage and semantic richness. Mirage's latent tokens, constrained by rigid priors, attend almost exclusively to weak local cues near the image borders throughout the inference process, leading to a restricted and fragmented exploration pattern. In contrast, CoLVR imposes no such strict constraints; its latent tokens learn more exploratory and discriminative features, yielding attention maps that distribute broadly across the global context. Consequently, CoLVR captures richer structural relationships and exhibits a markedly greater diversity in visual reasoning.

\subsection{Visualization of UMAP for a Single Case with Latent Tokens Rolled Out K Times}
To further investigate the exploratory capacity of latent tokens, we conduct an additional quantitative analysis at the level of individual cases. Specifically, we randomly sample 40 cases from the evaluation set. For each case, both Mirage and \ours{} are executed 20 times to generate multiple latent trajectories under stochastic conditions. We then employ UMAP to visualize the distribution of the resulting latent tokens and quantify their overall clustering behavior.

\begin{figure}[h]
  \centering
  \includegraphics[width=0.5\linewidth]{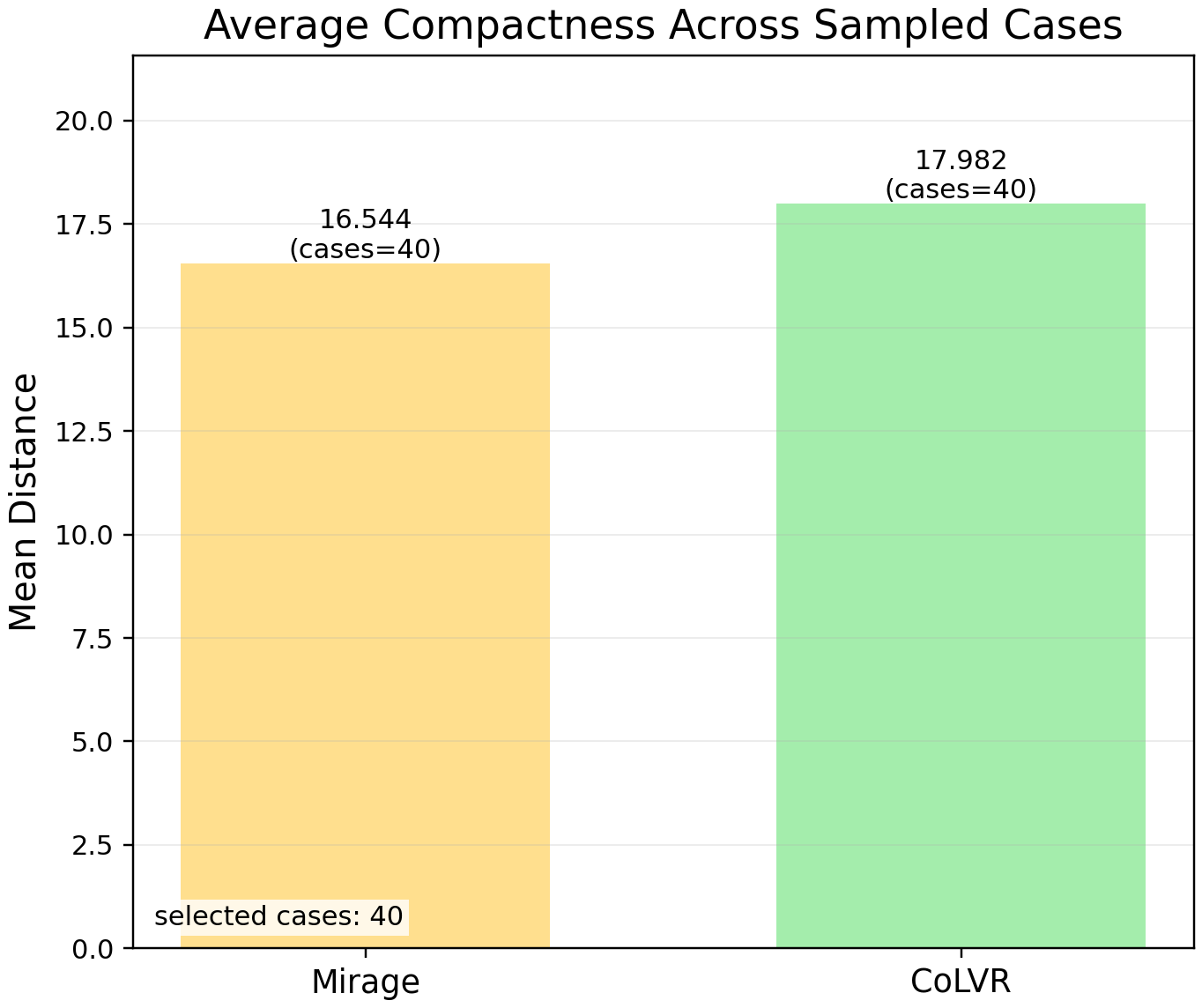}
  \caption{\textbf{Comparison of latent token dispersion across models}, measured by average distance to cluster centroids over 40 cases with 20 samples each.}
  \label{fig:appendix-umap}
\end{figure}

Concretely, for each model and each case, we compute the average distance from all latent tokens to their corresponding cluster centroid, which serves as a proxy for the degree of semantic dispersion. A larger average distance indicates stronger exploration in the latent space, while a smaller value suggests more concentrated representations. Finally, we average this metric across all sampled cases to obtain a global quantitative indicator of semantic exploration for each model. As shown in Fig.~\ref{fig:appendix-umap}, CoLVR achieves an average clustering metric of 17.982, while Mirage reports a lower value of 16.544. This difference suggests that the latent tokens produced by CoLVR are more dispersed in the latent space, indicating a stronger ability to explore a wider range of semantic representations. Such behavior effectively expands the solution space and allows the model to capture alternative reasoning paths. By comparison, the lower clustering metric of Mirage indicates that its latent representations are more concentrated, which points to more limited exploration. This pattern is consistent with its use of stricter constraints, which restrict the evolution of latent tokens and reduce the diversity of semantic trajectories.

\subsection{Attention Map}
\begin{figure}[t]
  \centering
  \includegraphics[width=1.0\linewidth]{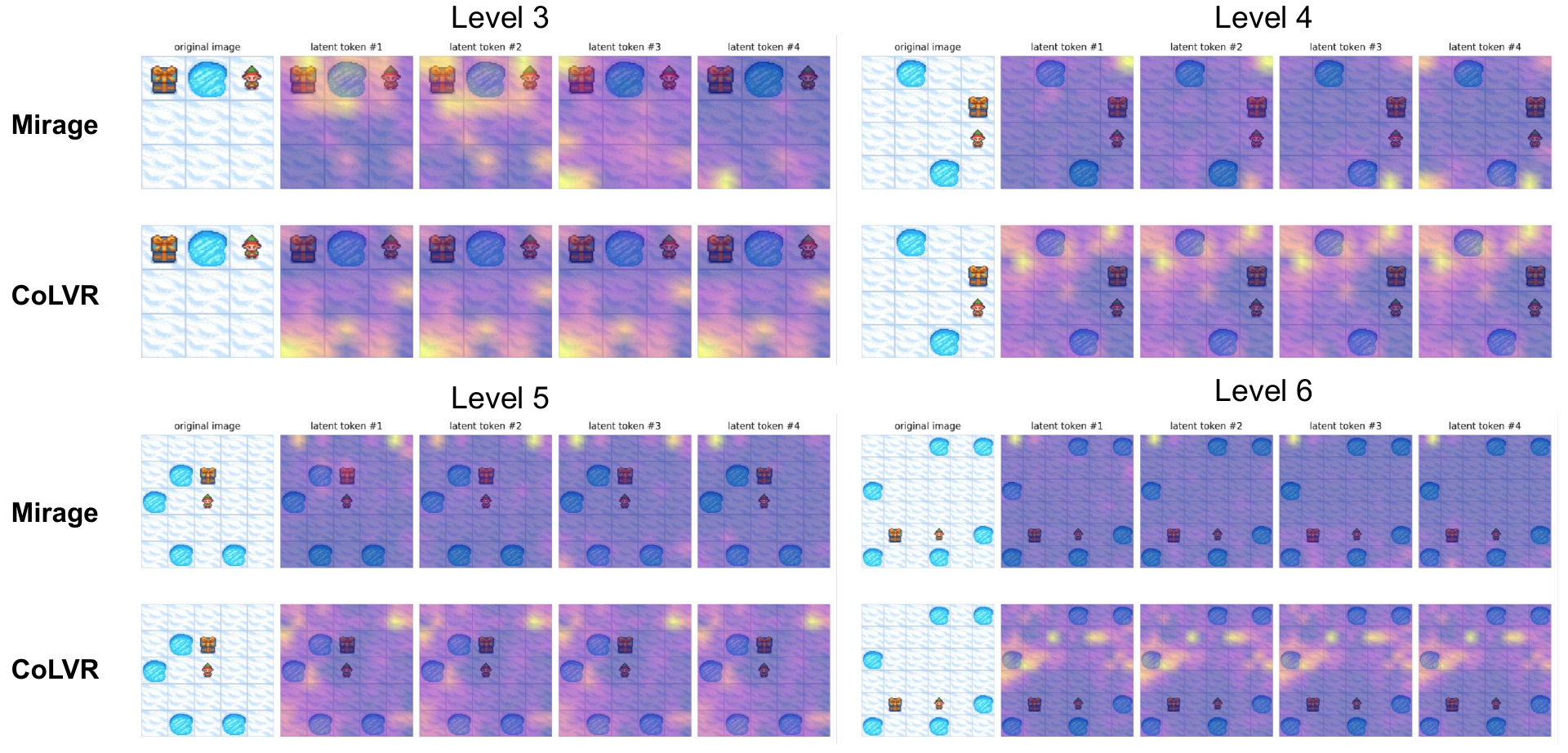}
  \caption{\textbf{Comparison of attention maps} across VSP levels for Mirage and CoLVR.}
  \label{fig:appendix_attenmap}
\end{figure}

Figure~\ref{fig:appendix_attenmap} compares the attention maps of Mirage and CoLVR across different VSP levels. As the reasoning process progresses, Mirage exhibits highly concentrated and repetitive attention patterns, where the activated regions remain relatively fixed across latent reasoning steps. This behavior suggests limited exploration of the visual space and indicates that the latent tokens tend to collapse into similar attention distributions. In contrast, CoLVR consistently produces more diverse and spatially dynamic attention maps. The activated regions shift across different latent tokens and cover a broader range of informative areas in the image, especially around task-relevant objects and surrounding contextual regions. This phenomenon is particularly evident in more complex levels such as Level~5 and Level~6, where CoLVR maintains exploratory attention behaviors while Mirage becomes increasingly static. These results demonstrate that CoLVR encourages more exploratory latent visual reasoning and enables the model to interact with visual content in a more flexible and comprehensive manner.

\subsection{Task Details}
\begin{figure}[h]
  \centering
  \includegraphics[width=0.7\linewidth]{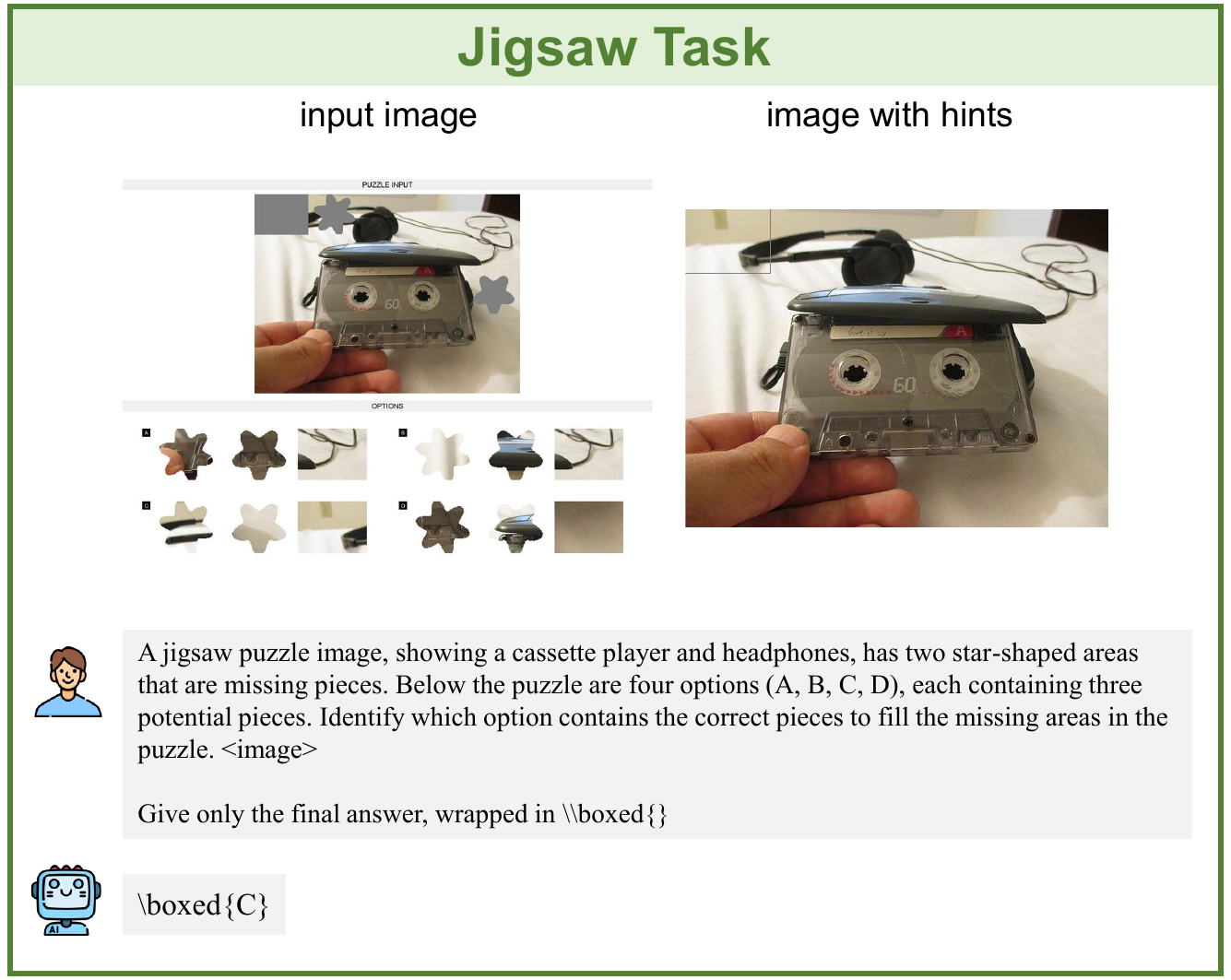}
  \caption{\textbf{Inference example: Jigsaw}.}
  \label{fig:append_jigsaw_case}
\end{figure}

\begin{figure}[h]
  \centering
  \includegraphics[width=0.7\linewidth]{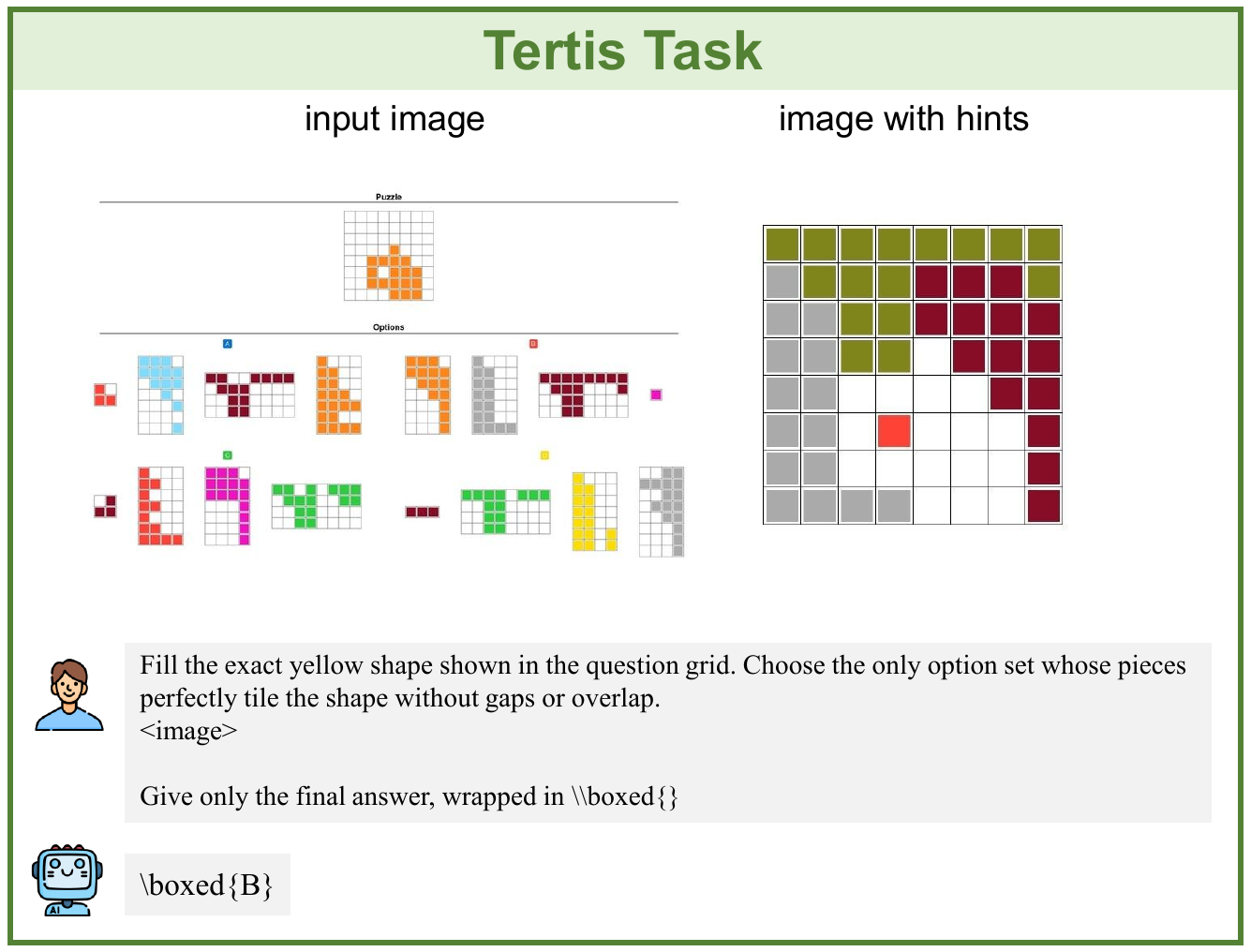}
  \caption{\textbf{Inference example: Tertis}.}
  \label{fig:append_tertis_case}
\end{figure}

\begin{figure}[h]
  \centering
  \includegraphics[width=0.7\linewidth]{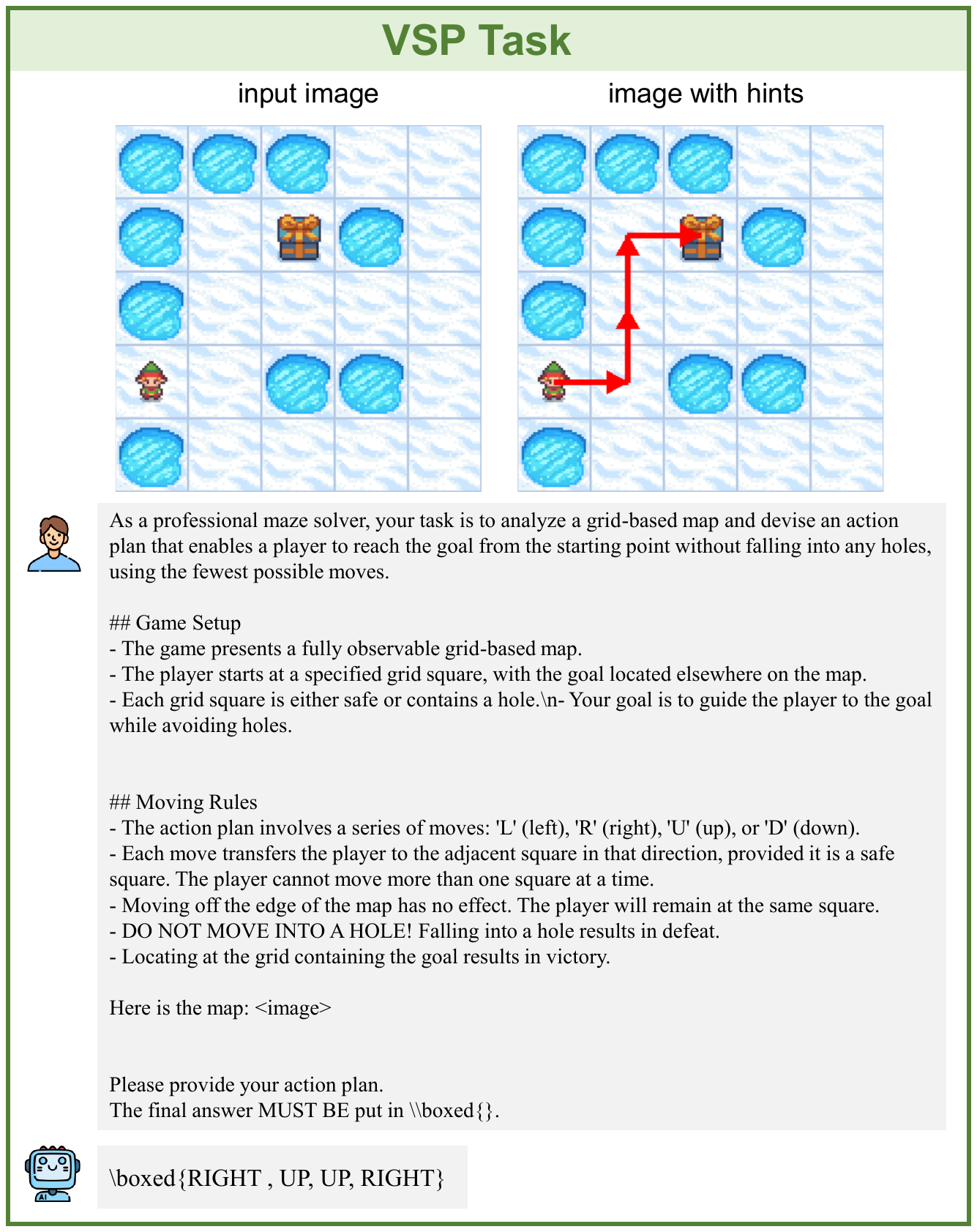}
  \caption{\textbf{Inference example: VSP}.}
  \label{fig:append_vsp_case}
\end{figure}

Here we introduce specific examples corresponding to our training and evaluation tasks, including input images and image examples with hints (as shown in Figure~\ref{fig:append_jigsaw_case}, ~\ref{fig:append_tertis_case}, ~\ref{fig:append_vsp_case}).

\begin{figure}[h]
  \centering
  \includegraphics[width=1.0\linewidth]{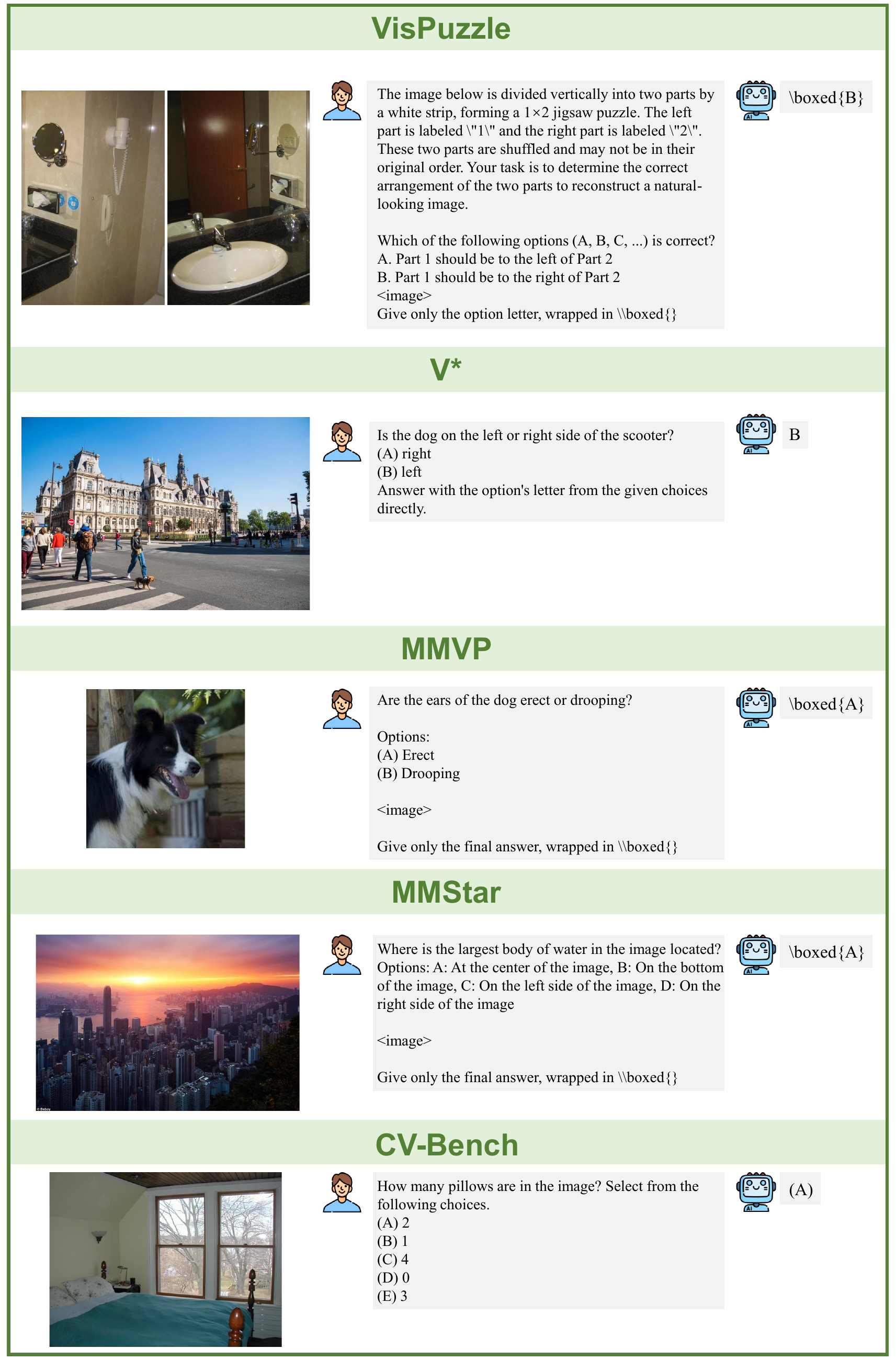}
  \caption{\textbf{Inference examples: VisPuzzle, V*, MMVP, MMStar, and CV-Bench}.}
  \label{fig:append_ood_case}
\end{figure}

Additionally, we provide illustrative examples from a range of out-of-domain benchmark tasks, including VisPuzzle, V*, MMVP, MMStar, and CV-Bench. These examples are intended to highlight the diversity and complexity of evaluation scenarios beyond the training distribution, offering a more comprehensive view of model generalization across varied visual reasoning and multimodal understanding tasks (As shown in Figure~\ref{fig:append_ood_case}).


\end{document}